# ΩSFormer: Dual-Modal Ω-like Super-Resolution Transformer Network for Cross-scale and High-accuracy Terraced Field Vectorization Extraction

Chang Li [a], Yu Wang [a], Ce Zhang [b] and Yongjun Zhang [c]

[a] *Key Laboratory for Geographical Process Analysis & Simulation of Hubei Province, and College of Urban and Environmental Science, Central China Normal University, Wuhan, China;*

[b] *Industry and Information Technology Bureau of Lingbao City, Henan Province, China, 472500;*

[c] *School of Remote Sensing and Information Engineering, Wuhan University, China.*

**ABSTRACT**

Terraced field is a significant engineering practice for soil and water conservation (SWC). Besides ecological benefits, it could increase soil organic matter content, promote carbon fixation and reduce greenhouse gas emissions, which is conducive to accelerating the realization of the "dual-carbon" objective (i.e., carbon peak and carbon neutrality). Terraced field extraction from remotely sensed imagery is the foundation for monitoring and evaluating SWC, whose research is worth paying attention to the following aspects: (1) there is currently no dual-modal (e.g., remotely sensed imagery and DEM) raster-vector dataset (DMRVD) for deep-learning-based terraced field vectorization extraction (TFVE); (2) downsampling in the encoder step of classical semantic segmentation networks leads to poor edge segmentation accuracy; (3) it is necessary to deeply investigate the effect of single or double-branch input method for improving extraction accuracy; (4) the dual-modal input method cannot guarantee a unified resolution or scale, so cross-scale effects remain to be explored; (5) the current pixel-level extraction of remotely sensed imagery is almost a raster result, which does not meet the practical application requirements for lightweight and infinite scalability. To address the issues above, this study is the first to propose a novel dual-modal Ω-like super-resolution Transformer network (ΩSFormer) for intelligent TFVE, offering the following advantages: (1) reducing edge segmentation error from conventional multi-scale downsampling encoder, through fusing original high-resolution features with downsampling features at each step of encoder and leveraging a multi-head attention mechanism; (2) improving the accuracy of TFVE by proposing a Ω-like network structure, i.e., dual-modal and dual-branch ($D^2MB$), which fully integrates rich high-level features from both spectral and terrain data to form cross-scale super-resolution features; (3) validating an optimal fusion scheme for cross-modal and cross-scale (i.e., inconsistent spatial resolution between remotely sensed imagery and DEM) super-resolution feature extraction; (4) mitigating uncertainty between segmentation edge pixels by a coarse-to-fine and spatial topological semantic relationship optimization (STSRO) segmentation strategy; (5) leveraging contour vibration neural network to continuously optimize parameters and iteratively vectorize terraced fields from semantic segmentation results. Moreover, a DMRVD for deep-learning-based TFVE was created for the first time, which covers nine study areas in four provinces of China, with a total coverage area of 22441 km². Those regions cover different terrains and landforms, with various geographical conditions and significant spatial differences. DMRVD with 19330 remotely sensed imageries, 19330 DEMs and 371150 vectorized polygons. To assess the performance of ΩSFormer, classic and state-of-the-art (SOTA) networks were compared with U-Net, U-Net++, Deeplabv3+, HRNet, W-Net, W-Net++, HRFormer, Segformer, and SAM etc. For cross-scale dual-modal datasets, ΩSFormer achieved the best performance with mIOU and OA values of 0.976 and 0.981 respectively. The mIOU of ΩSFormer has improved by 0.165, 0.297 and 0.128 respectively, when compared with best accuracy single-modal remotely sensed imagery, single-modal DEM and dual-modal result. DMRVD and ΩSFormer code are available at https://github.com/ccnuer-yw/DMRVD.git.

## 1. Introduction

Soil water erosion represents a significant threat to land resources and the ecological environment (Duan et al. 2020), impeding sustainable economic development. Soil water erosion compromises the hydrological function of the soil and reduces food production (Coban et al. 2022). It can also lead to the siltation of rivers, reduce the effectiveness of water conservancy projects pollute water quality, endanger ecosystems and the safety of human drinking water; severe soil water erosion significantly impacts the capacity to sequester carbon, which in turn contributes to global warming and hinders the achievement of dual carbon objectives, including carbon peaking and carbon neutrality. With soil water erosion currently surpassing the capacity for regeneration, thereby emerging as a global environmental challenge (Amundson et al. 2015). Therefore, soil and water conservation (SWC) is essential. In the Chinese Soil Loss Equation (CSLE), tillage practices ($T$) is an import SWC factor (Baoyuan et al. 2002). As a tillage practices, terraced field provides various ecosystem services (Lu et al. 2023), agricultural applications (Weiss et al. 2020), and effectively prevents water flow from eroding the soil (Serbin et al. 2009). Furthermore, farmlands (e.g., terraced fields) contribute to achieving carbon peaking and carbon neutrality (Lin et al. 2022). In light of the increasing importance of protecting and monitoring terraced fields, remote sensing monitoring and assessment have emerged as crucial tools for intelligently extracting terraced field. However, classical methods suffer from certain limitations. With the continuous development and widespread application of AI, deep-learning-based extraction of terraced fields from remotely sensed imagery holds significant research and application potential.

Recently, semantic segmentation of remotely sensed imagery has been widely applied in several fields (He et al. 2024; Wieland et al. 2023). In the field of terraced field extraction, studies based on traditional machine learning include the following: (Luo et al. 2020; Pijl et al. 2020; Zhang et al. 2017; Zhao et al. 2017). However, with the contemporary machine learning developing, deep-learning-based terraced field extraction has been studied, of which the only literature reports are as follows: Zhao et al. (2023) introduced a strip pooling module and an attention mechanism based on the DeepLabV3+, to achieve high-accuracy extraction of terraced field from UAV imagery; Zhang et al. (2024) improved the DeepLabV3+ and combined it with transfer learning to enhance the accuracy of terraced field extraction in high-resolution remotely sensed imagery; Peng et al. (2024) utilized an improved U-Net combined with DEM to achieve automatic extraction of terraced field in remotely sensed imagery of hilly areas. Zhao et al. (2024) introduced ANB-LN and DFB modules to form the NLDF-Net framework for extracting terraced fields from remotely sensed imagery.

The essence of the aforementioned applied research is deep-learning-based method and techniques for semantic segmentation, including the following research direction: (1) Data-driven datasets. Effective semantic segmentation in remote sensing requires large-scale and diverse remotely sensed imagery datasets for the training and evaluation of deep learning networks (DNN). With the rise of deep learning in remote sensing, numerous datasets have been developed for the semantic segmentation of remotely sensed imagery, such as ISPRSPostdam, ISPRSVaihingen, Inria Aerial Image Labeling Dataset (Maggiori et al. 2017), DeepGlobe 2018 (Demir et al. 2018), UAVid (Lyu et al. 2020), GID (Tong et al. 2020), LandCover.ai (Boguszewski et al. 2021), UrbanBIS (Yang et al. 2023), CASID (Liu et al. 2023a), OpenEarthMap (Xia et al. 2023), CASID (Liu et al. 2023a). In the context of semantic segmentation, previous works have not yet addressed the issue of dual-modal remote sensing datasets for deep-learning-based terraced field extraction. Therefore, it is urgently necessary to construct a remotely sensed imagery dataset for terraced field extraction and SWC applications.

(2) Network models. As a pivotal task in the field of remote sensing interpretation, semantic segmentation has gradually evolved from traditional machine learning to deep learning, which includes the following network models:

① Convolutional neural network (CNN) is designed to process and analyze visual data by automatically learning spatial hierarchies of features through convolutional layers. It has had a profound impact on the field of semantic segmentation. The classical CNN semantic segmentation networks, U-Net (Ronneberger et al. 2015), U-Net++ (Zhou et al. 2020b), ResNet (He et al. 2016a, b), DeepLab (Chen et al. 2018b) and DeepLabV3+ (Chen et al. 2018a) demonstrated efficacy in image semantic segmentation. Zhang et al. (2021b) employed CNN for hyperspectral remotely sensed imagery classification and achieved satisfactory results. HRNet used CNN as a core component of its architecture to process and maintain high-resolution feature representations across multiple scales (Wang et al. 2021). FarSeg++ proposed a foreground-aware network for object segmentation in high-resolution remotely sensed imagery, effectively enhancing the extraction capabilities of CNN (Zheng et al. 2023).

② Transformer uses self-attention mechanisms to process and generate sequences, enabling efficient handling of long-range dependencies in data. It has been gradually integrated into semantic segmentation. For instance, SegFormer combined Transformers with lightweight multi-layer perceptron (MLP) decoders (Xie et al. 2021). HRFormer utilized a high-resolution convolutional network and employed local window self-attention (Yuan et al. 2021a). Swin Transformer introduced the hierarchical design and the shifted windows method to enable self-attention computation at different scales, thereby enhancing semantic segmentation outcomes (Liu et al. 2021). TransUNet utilized U-Net and applied a self-attention mechanism to feature extraction and pixel-level classification tasks (Chen et al. 2021). EfficientViT (Liu et al. 2023b), segment anything model (SAM) (Kirillov et al. 2023) and EfficientSAM (Xiong et al. 2024) all achieved remarkably high performance on semantic segmentation.

③ Recurrent neural network (RNN) is designed to process sequential data by retaining information through loops in its hidden layers, enabling it to learn temporal patterns. It enables the incorporation of temporal information correlation into semantic segmentation. The FCN-RNN, which represented a classic approach to semantic segmentation, introduced an RNN structure to facilitate the processing of temporal information and enhance performance in remotely sensed imagery interpretation and other fields (Visin et al. 2016). Other RNN networks also demonstrated the full utilization of their ability to capture temporal information and possess spatial correlation structures, achieving satisfactory results in semantic segmentation (Le et al. 2018).

④ Mamba is a deep learning architecture focused on sequence modeling for high-performance processing (Gu and Dao 2024). Especially, it combines the sequential modeling capabilities of RNNs with the parallelization advantages of Transformers. Through a simplified state space model, it achieved modeling capabilities comparable to Transformers while maintaining linear time complexity, which provided significant advantages in handling long sequences. Although Mamba has certain advantages in computational efficiency, the cross-attention mechanism of vision Transformer (ViT) offers much stronger context-capturing capabilities.

(3) Multi-modal-based semantic segmentation. In the field of computer vision, leveraging multimodal data (e.g., speech, text, image, etc.) for semantic segmentation has become a growing trend. However, in the field of remote sensing, different sensors (passive and active sensors, e.g., RGB imagery and LiDAR point clouds), different spectral bands (e.g., true-color and false-color imagery), and different data types (e.g., imagery and DEM/DSM) can also constitute multi-modal data. There are two approaches to dual-modal network inputs: one is single-branch input (Gao et al. 2024), and the other is dual-branch input (Deng et al. 2019; Vachmanus et al. 2021; Yang et al. 2021).Regardless of the input approach, the scale differences between the dual-modal inputs can impact the features that can be extracted. Existing research has begun to focus on the scale effects present in remotely sensed imagery. The application of multi-scale feature extraction in deep learning has led to an increased awareness of the scale effect in remotely sensed imagery (Ding et al. 2018; Jiang et al. 2021). Li et al. (2020) proposed an adaptable two-stream framework, employing low-resolution input to preserve high-resolution representations for segmentation. Liu et al. (2024) demonstrated that scale changes impact image semantic segmentation performance.

(4) Super-resolution semantic segmentation. Super-resolution technology is a classic technique in the field of image processing and has recently been introduced into the semantic segmentation of remotely sensed imagery. DSRL proposed a dual-stream framework that maintains high-resolution representations and reduces computational complexity under low-resolution inputs, thereby effectively improving the accuracy and efficiency of semantic segmentation (Li et al. 2020). Zhang et al. (2022) proposed a collaborative network capable of simultaneously handling both super-resolution and semantic segmentation tasks in remotely sensed imagery, achieving high-resolution semantic segmentation and super-resolution reconstruction results even with low-resolution input images. Aakerberg et al. (2022) utilized an auxiliary semantic segmentation network to guide super-resolution learning, enabling super-resolution processing of real-world low-resolution images without the need for high-resolution images as ground truth references.

The related research is worth paying attention to as follows:

(1) No multimodal or vectorization dataset for deep-learning-based terraced field extraction

In semantic segmentation, open-source datasets of agricultural scenes typically include various categories of farmland, while terraced fields have unique morphological structures and edge features that distinguish them from other land types. Currently, it seems that only Google Earth provides cost-free high spatial resolution remotely sensed imagery with only visible light, but there is no near-infrared band. However, the extraction of terraced field based on visible light RGB images presents a new challenge. Hence, adding DEM and combining it with visible light imagery to form multimodal complementary advantages is worth paying attention to. Although there are currently deep-learning-based terraced field extraction datasets available (Do et al. 2019; Yu et al. 2022; Zhang et al. 2024; Zhao et al. 2024), they are not open-source.

Furthermore, there have been no reports on dual-modal (e.g., remotely sensed imagery and DEM) raster-vector dataset (DMRVD) for deep-learning-based terraced field vectorization extraction (TFVE). Current pixel-level extraction techniques for remotely sensed imagery primarily generate raster data. While this approach has certain applicability in image analysis, it suffers from significant limitations in practical applications. Primarily, raster data suffers from inherent spatial resolution limitations, leading to aliasing effects that challenge high-accuracy and multiscale analysis requirements. Moreover, raster data often entails substantial data volumes, which not only escalate storage and transmission burdens but also markedly elevate computational complexity and resource requirements, posing challenges for lightweight applications. It is therefore imperative to develop vectorized extraction techniques for remotely sensed imagery to achieve infinite scalability and lightweight processing, in order to meet the diverse requirements of different application scenarios.

(2) No DNN for terraced field vectorized extraction

The majority of current researches employ target pixel extraction features to achieve semantic segmentation. However, due to the distinctive topographic characteristics of terraced field, including regular hierarchical structures and intricate boundary patterns, the raster results of pixel-level extraction are constrained in their ability to address terraced field extraction. Vectorized extraction offers a number of advantages, including the ability to provide high-accuracy and high-resolution terrain boundary information. Additionally, it enables the lightweight extraction of terraced field, which is crucial for addressing the infinite scalability problem in semantic segmentation. However, there is currently no vectorized DNN that is specifically designed for terraced field extraction. Consequently, there is a pressing need to investigate the potential of DNNs for the vectorized extraction of terraced fields. This approach is expected to significantly improve the accuracy and resolution of terraced field edge delineation, thereby meeting the demand for efficient and intelligent extraction of terraced fields.

(3) Input design of dual-modal network for terraced field extraction

There are variations in the input methods of dual-modal networks. Some approaches involve fusing dual-modal data into multi-channel images for single-branch input, while other methods separately input dual-modal data in a dual-branch mode. Although dual-modal input methods can enrich the available information and extract enhanced features, the different input methods of dual-modal (single-branch or multi-branch) can also affect semantic segmentation accuracy. This aspect is worth exploring, and its underlying mechanisms need to be elucidated.

(4) Semantic segmentation loss of edge accuracy

In order to reduce the computational burden and improve the efficiency of the model, downsampling is typically employed in the semantic segmentation encoder step. This inevitably results in the loss of detail information due to the reduction in spatial dimensions of the input feature maps. In the decoder step, upsampling is performed on the downsampling feature map in order to restore the spatial dimension of the original input. Nevertheless, this process does not fully restore the abstracted details, particularly in the edge regions. Due to the restricted sampling rate, it is not feasible to accurately restore each pixel of the original feature map, which results in the loss of further information at the edges.

(5) Value of spatial topological semantic relationship

Spatial topological semantic relationship provides crucial information about the location and interrelationships between geographic objects, which is essential for understanding the geospatial semantic structure. By considering the spatial topological relationships between geographic objects, it is possible to more accurately distinguish and recognize land cover types that have similar spectral characteristics but different spatial distributions. In terraced field extraction, there are specific spatial topological relationships (e.g., containment, adjacency, and intersection) and spatial semantic attributes (e.g., label for land cover type) between geographic objects. These relationships are systematic and serve as key criteria for extraction. As an auxiliary method for optimizing semantic segmentation accuracy, spatial topological semantic relationship can enhance ability of the model to capture global spatial.

(6) The effect of cross-modal and cross-scale with super-resolution

As the utilization of multi-source remotely sensed imagery becomes increasingly prevalent, this trend also gives rise to the issue of scale heterogeneity. The spatial heterogeneity of the second law of geography pertains to the fact that observations are frequently obtained at disparate scales. In the context of multi-source images of varying resolutions, research in deep-learning-based super-resolution semantic segmentation is confronted with a series of challenges and limitations. The development of multi-source data and the application of dual-modal DNNs has highlighted a gap in research on the scale effects caused by scale differences between dual-modal. Consequently, the effective utilization of diverse dual-modal data to extract more comprehensive feature information remains a pressing issue in contemporary research. Therefore, the cross-modal scale effects and super-resolution fusion issues associated with semantic segmentation of multi-source remotely sensed imagery with different spatial resolutions require further investigation.

This paper has the following advantages and contributions:

(1) A new dataset (i.e., DMRVD) for TFVE is established for the first time.

TFVE requires high-resolution data that accurately captures the spectral and topographic characteristics of actual terraced field. Although there are existing remote sensing datasets that include terraced field categories, to the best of our knowledge, there is currently no dataset specifically designed for the intelligent vectorized extraction of terraced fields. Moreover, there is an absence of dual-modal deep-learning-based terraced field extraction datasets. This study is the first to present a deep-learning-based DMRVD for super-resolution-fusion-based TFVE. The DMRVD encompasses nine counties, exhibiting notable spatial disparities and a low degree of spatial correlation. This dataset provides cross-modal and cross-scale super-resolution features, enabling advanced applications in TFVE.

(2) A dual-modal Ω-like super-resolution Transformer network (ΩSFormer) is the first to be proposed for semantic segmentation of remotely sensed imagery.

In contrast to existing dual-modal DNNs that utilize single-branch input methods, this study introduces ΩSFormer, the first model specifically designed for super-resolution-fusion-based intelligent vectorization extraction. Due to the $D^2MB$ input method and the mechanism of continuously fusing original features with encoded (downsampled) features, ΩSFormer forms an Ω-like network architecture that is wide at the top and narrow at the bottom. ΩSFormer has the following advantages:

① Fully fusing dual-modal information; compared to the dual-modal single-branch input method, the $D^2MB$ input method, which fuses cross-modal and cross-scale super-resolution features of spectral and surface morphology data, yields a more comprehensive and sophisticated semantic understanding. Furthermore, the integration of cross-modal and cross-scale super-resolution feature markedly enhances the accuracy of semantic segmentation;

② Effectively improving edge segmentation accuracy; ΩSFormer fuses original high-resolution features at each step of encoder and employs a MHSA to enhance spatial resolution information during the encoder phase. ΩSFormer mitigates the issue of information loss during downsampling that cannot be fully recovered during decoder, thereby effectively improving the accuracy of semantic segmentation at the edges;

③ Obtaining more comprehensive spatial topological semantic relationship; ΩSFormer adopts a coarse-to-fine and spatial topological semantic relationship optimization (STSRO) segmentation strategy. This approach fully considers spatial dependencies to better understand the spatial distribution characteristics and semantic information of pixels, thereby reducing the uncertainty between semantic segmentation edge pixels;

④ Achieving lightweight and detailed edge contour expression, i.e., vectorization; contour vibration neural network is employed for vectorizing terraced fields, effectively mitigating the jagged artifacts commonly associated with raster images. Vectorized extraction takes full advantage of infinite scalability and makes terraced field edges clearer. It also facilitates data compression and supports complex spatial analysis.

(3) Cross-modal scale effect in super-resolution semantic segmentation is the first to be investigated.

Currently, an increasing number of sensors are available for acquiring remotely sensed imagery, which produce images with multiple resolutions. However, there is a limited availability of dual-modal data at the same resolution. To fully consider the impact of cross-modal scale effects, dual-modal images of varying resolutions are resampled to create datasets with cross-scaled, high-resolution and low-resolution images all at the same scale. This study establishes spatial relationships between digital elevation DEMs and remotely sensed imagery, utilizing information such as object edges and textures from high-resolution remotely sensed imagery to enhance the details of low-resolution DEMs. This approach improves label precision and accuracy, thereby achieving the objective of super-resolution fusion. Comparative experiments are conducted to analyze the scale effects arising from differences between cross-scale data.

To systematically compare the scale effects associated with cross-modal multisource remotely sensed imagery in semantic segmentation. Additionally, we aim to develop and validate an intelligent extraction method for optimally fusing cross-modal, cross-scalar super-resolution data.

(4) The effectiveness and mechanism of ΩSFormer for TFVE are analyzed.

Compared with previous TFVE studies, the mechanism of terraced field intelligent extraction of high-accuracy in dual-modal is analyzed based on ΩSFormer for the first time. By analyzing the underlying principles and key semantic segmentation strategies, the advantages of ΩSFormer have been revealed. By analyzing the unique $D^2MB$ input method and the super-resolution Transformer connection module (SRTCM) of ΩSFormer, this study conducts a pioneering mechanistic analysis of edge enhancement in deep-learning-based terraced field extraction for the first time. The effectiveness of ΩSFormer in the extraction task is confirmed by conducting comparative experiments with existing studies. Furthermore, this study demonstrates the important role of its individual modules in improving the accuracy of TFVE through ablation experiments.

## 2 Study area and datasets creating

### 2.1. Study areas

Terraced fields exhibit unique textures, spectral signatures, shapes and other characteristics influenced by their environment and geographical location. They can be classified into four types: level terraced fields, sloped terraced fields, reverse slope bench terraced fields, and slope-separated terraced fields. Different categories of terraced field have different construction plans and uses.

(1) Terraced fields are cultivated farmlands built along contour lines on sloping land, with stepped or sloping cross-sections (Arnáez et al. 2015). From the perspective of climate, terrain, and other factors, there are relatively few terraced fields in many continents and countries. For example, Russia, which has the largest area, is mainly covered in ice and snow all year round, and the USA is dominated by plains, so there are also almost no terraced fields. In contrast, the topography and climate of Asia are more conducive to the formation and distribution of terraced fields. China in particular, with its mountainous terrain and favorable climate, has developed a widespread distribution of terraced fields, which are typical and globally representative of terraced fields (Wei et al. 2016).

(2) DMRVD contain nine counties in southwestern China, where terraced fields are

most prevalent and the terrain varies greatly (Cao et al. 2021), with a maximum elevation difference of approximately 3,630 meters.

To ensure the representativeness of the data and to enhance the generalization and robustness of the model, nine regions in China were selected as study areas: Tonglu county in Zhejiang province, Qiaojia county, Ludian county in Yunnan province, Wudu district, Xihe county of Gansu province, Wangcang county, Yanting county, Cangxi county and Jiuzhaigou county of Sichuan province (Fig. 1).

Our study areas span different latitudes, have different climate types, and have varying natural geographic conditions and surface morphology, resulting in increased variability. These regions contain a rich variety of terraced fields with weak spatial correlation (i.e., significant spatial heterogeneity), which can be used to verify the generalizability of this research. The areas of Tonglu county, Qiaojia county, Ludian county, Wudu district, Xihe county, Wangcang county, Yanting county, Cangxi county, and Jiuzhaigou are 1829 km$^2$, 3245 km$^2$, 1484 km$^2$, 4511 km$^2$, 2398 km$^2$, 2977 km$^2$, 1,635 km$^2$, 2,337 km$^2$, 720 km$^2$, respectively. And their elevation differences are approximately 1196 meters, 3524 meters, 2,788 meters, 3496 meters, 1563 meters, 1900 meters, 400 meters, 1650 meters and 2764 meters, respectively. Therefore, they are representative of terraced fields.

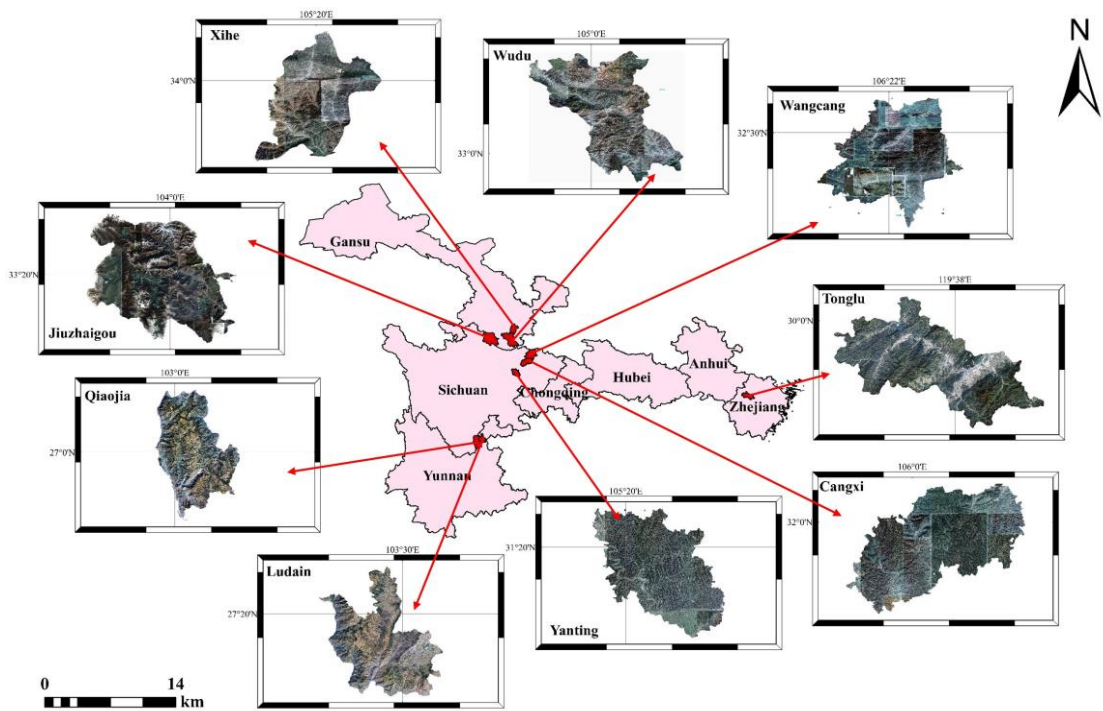

**Fig. 1.** Overview of the study areas.
**2.2. Dataset preprocessing and creating**

Google Earth integrates high-resolution remotely sensed satellite imagery and vector data from multiple sources. In this study, remotely sensed imagery with a resolution of 2 meters and DEM with a resolution of 12.5 meters are acquired using Google Earth in 2022. The acquisition of optical images is invariably accompanied by cloud cover, which introduces varying degrees of obscuration of the ground information. This phenomenon significantly impacts the overall quality of the image. Therefore, this study prioritizes the use of cloud-free data for experimentation whenever feasible. In the context of geographic big data, the construction of a dataset can only provide accurate prior knowledge and spatial information after rigorous processing. Dataset for validation was obtained after the following processing steps:

(1) Image calibration, encompassing both geometric and radiometric corrections based on stringent criteria, is systematically performed. Subsequently, the remotely sensed imagery and DEM are aligned within the same coordinate system.

(2) Image registration, involving the alignment of remotely sensed imagery to a consistent coordinate system using DEM as a reference. This process was semi-automated and involved selecting a sufficient number of control points that were uniformly distributed across the image. A polynomial transformation model was employed to match remotely sensed imagery and DEM at the pixel level. During this matching process, the spatial correspondence between the remotely sensed imagery and the DEM was optimized by adjusting the transformation parameters, thereby enhancing the accuracy and stability of matching process.

(3) Label production, utilizing the vectorization of high-resolution remotely sensed imagery over terraced field areas, aims to generate vector datasets for training sample regions. Due to the distinctive dual-modal structure of this study, remotely sensed imagery is overlaid onto the 3D model space for annotation, forming the basis for visual interpretation. To precisely delineate object boundaries, eight key interpretation elements: size, shape, shadow, color, texture, pattern, location, and contextual relationships are leveraged. Building on visual interpretation, these elements are integrated into an automated annotation process for target features. To ensure annotation accuracy, various band combinations are employed to enhance feature differentiation and improve interpretability. Vector data not only accurately delineates the contours and shapes of objects, leading to improved accuracy and more detailed semantic segmentation results, but also facilitates rapid object detection, boundary tracking, and region partitioning. Finally, the obtained vector data is rasterized to form the raster label data used for training the DNN. The utilization of DMRVD can enhance the efficiency of semantic segmentation while simultaneously ensuring optimal extraction accuracy.

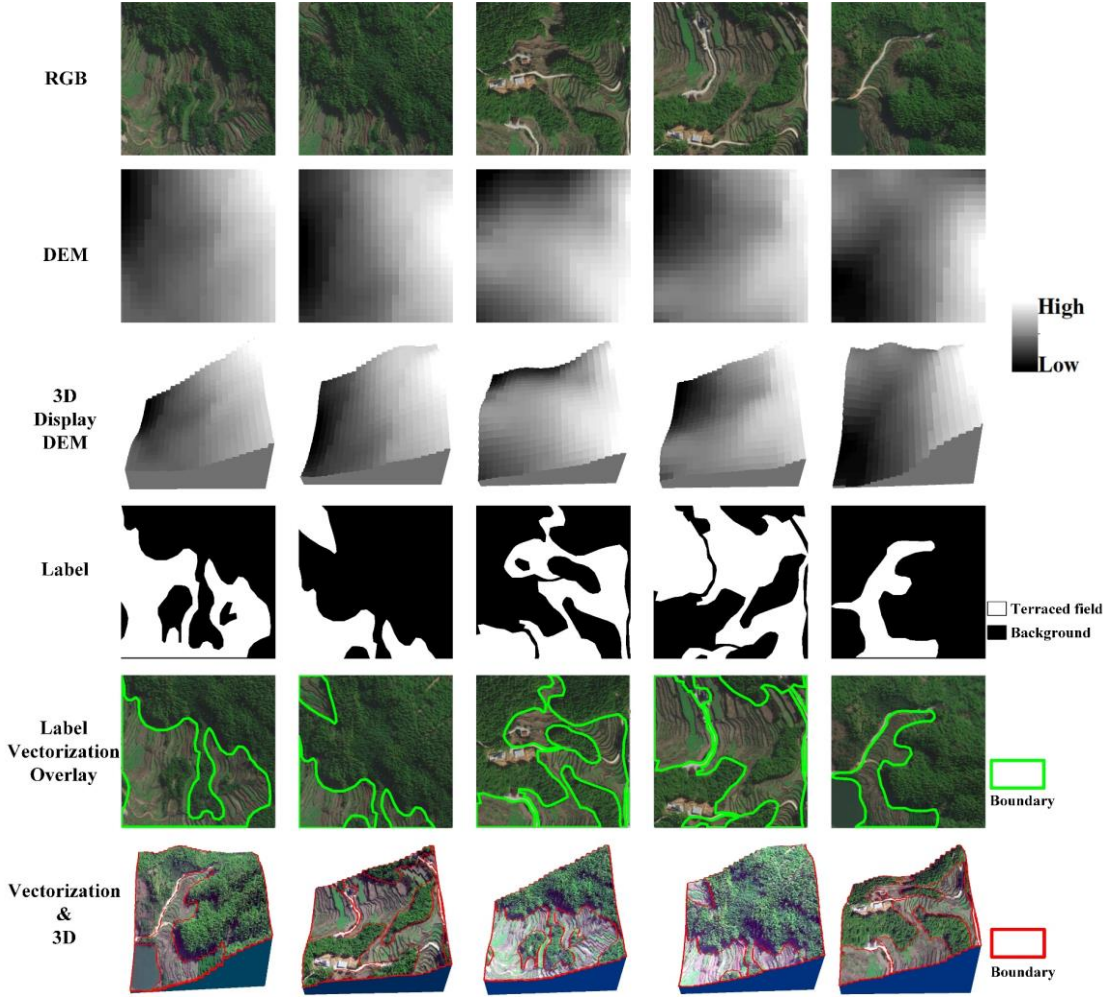

**Fig. 2.** DMRVD display. The first row is remotely sensed imagery, the second row is DEM, the third row is 3D display of DEM, and the fourth row is binarized label. The last row is label vectorization display of overlaying 3D.

(4) Image cropping, segmenting the entire scene into smaller images using a sliding window at certain intervals. We collected remotely sensed imageries and DEMs, each sized 512×512 pixels, to serve as the source domain for ΩSFormer.

(5) Dataset screening and enhancement, examining the images in a 1:1 ratio between terraced fields and non-terraced fields, resulting in the acquisition of a total of 19330 images and 19330 DEMs. DMRVD is divided into train, validation and test sets with the ratio of 8:1:1. In this study, image rotation, misalignment transformation and brightness adjustment were utilized to enhance the samples.

In order to ascertain the optimal scale fusion scheme under cross-scale conditions (i.e., inconsistent spatial resolution between remotely sensed imagery and DEM), This study processes the remotely sensed imageries, labels, and DEMs to create DMRVD with both uniform and varying resolutions, providing data for comparative experiments. The specific experimental groupings are shown in Table 1.

Table 1. The images resolution of DMRVD.

| grouping | | remotely sensed imagery | DEM | label |
| --- | --- | --- | --- | --- |
| cross-scale | (a) | 2m | 12.5m | 2m |
| high-resolution same-scale | (b) | 2m | 2m | 2m |
| low-resolution same-scale | (c) | 12.5m | 12.5m | 12.5m |

## 3 Methodology

Firstly, remotely sensed imagery and DEM are input into SRTCM in a dual-stream, then outputs soft object regions. Its advantage lies in proposing super-resolution multi-scale techniques to enhance the information available during encoder. By fusing the original high-resolution features with downsampling features at each step of encoder, SRTCM alleviates the issue of detailed features being lost during decoder. SRTCM proposed to extract spectral information of images and surface morphology features of DEM separately from the bimodal dataset using $D^2MB$ input method. SRTCM captures advanced and rich integrated features, thereby enhancing the accuracy of edge extraction.

Subsequently, soft object regions are input into STSRO, then outputs semantic segmentation results. STSRO leverages spatial topological semantic relationship to enhance the accuracy of semantic segmentation. This is achieved by characterizing a pixel in accordance with the category of the corresponding object, thereby augmenting the pixel representation.

Ultimately, the binarization results of the semantic segmentation are input into vectorization extraction module (VEM), then outputs object boundary contours. The advantage of VEM lies in utilizing a vibration model from the image feature domain to dynamically optimize edge parameters, which allows for further processing of the segmentation results to achieve vectorized extraction of edge contours. VEM is capable of performing complex spatial analysis while simultaneously enabling data compression and efficient edge vectorization extraction.

### 3.1. Cross-modal cross-scale interactions of input data

Due to limitations in data access, remotely sensed imagery and DEM are available at varying resolutions. Semantic segmentation is a task that involves predicting dense pixels, making spatial resolution crucial. Discrepancies across-scale between multi-modalities can lead to scale effects. To evaluate these cross-modal scale effects in the TFVE, the data must be preprocessed before being input into ΩSFormer. In this article, DMRVD is resampled to create three testing comparisons for investigating the cross-modal scale effects, as detailed in Table 1.

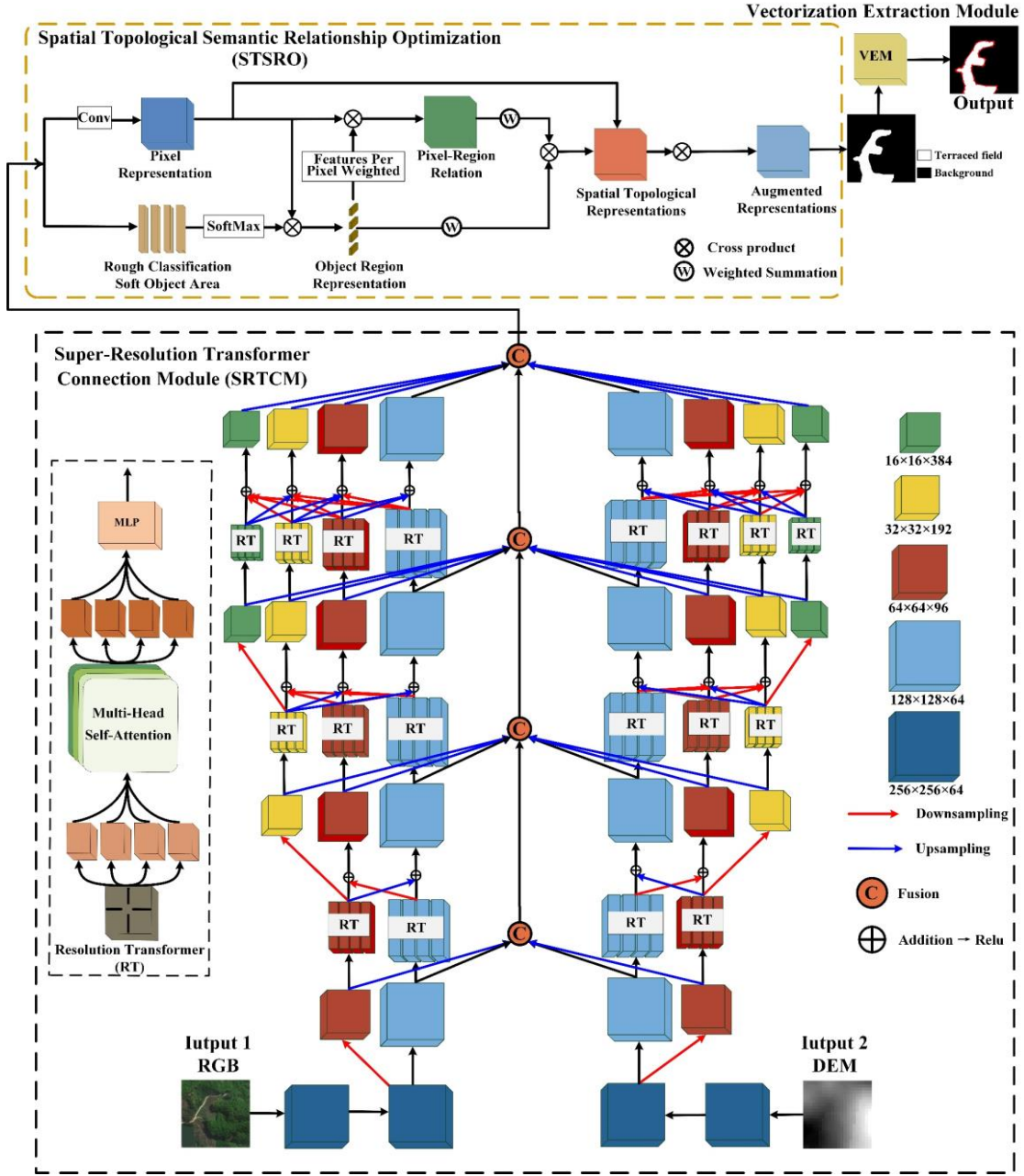

**Fig. 3.** Structure of the proposed ΩSFormer. ΩSFormer continuously fuses low-resolution features with high-resolution features at each step of encoder, wherein the more to the upper layer of the code the more scale features are fused, so it constitutes

the structure of the upper wide and lower narrow. The D²MB input method is arranged in a symmetric configuration, thereby forming an Ω-like network structure. The black box represents the feature extraction and feature fusion component, while the yellow box encompasses STSRO, which performs additional computation on the extracted features to yield the final semantic segmentation result. VEM is vectorization extraction module, utilizes the binary segmentation results of STSRO to extract the contour of terraced fields.

**3.2. Super-resolution Transformer connection module**

The first stage of ΩSFormer starts with a high-resolution convolutional backbone and keeps that high-resolution feature always. Subsequently, high-to-low resolution branches generated through downsampling are incrementally incorporated as a new stage in the convolutional process, where multiple features at varying scales are convolved separately in a parallel fashion. Consequently, ΩSFormer is capable of maintaining high resolution without resorting to a recovery of the feature map resolution in a manner that transitions from low resolution to high resolution. The SRTCM of ΩSFormer differs from the existing encoder-decoder structure. This distinction effectively alleviates the issue of information loss caused by encoder downsampling that cannot be fully recovered during decoder. Therefore, SRTCM facilitates the efficient preservation of an accurate spatial representation of the feature map and enhances the accuracy of edge segmentation.

The core element of ΩSFormer comprises a series of stages, each meticulously crafted to enhance feature representations across varying resolutions. Within each stage, the feature representations of individual resolution streams are iteratively refined through a series of independent Transformer blocks. Simultaneously, cross-resolution information exchange is enabled by a meticulously designed convolutional multi-scale fusion module. This innovative framework guarantees that the output from each scale branch reflects a cohesive fusion of results from all branches, thereby enabling the effective integration of attention mechanisms that capture both short- and long-range spatial dependencies.

ΩSFormer assigns feature maps $\mathbf{X} \in \mathbb{R}^{R \times C}$ to a series of small windows that do not overlap each other, each of size $K \times K$. Within each window, the MHSA (Huang et al. 2019) is applied independently, incorporating relative position information into elf-attention of the local window (Raffel et al. 2020). However, this design raises concerns about the potential difficulty in effectively communicating information across windows. The formula for dividing the small window is as follows:

$$\mathbf{X} \rightarrow \{\mathbf{X}_1, \mathbf{X}_2, \mathbf{X}_3, \ldots, \mathbf{X}_n\} \tag{1}$$

Where **X** is feature map and $\mathbf{X}_n$ is the feature of the *n*-th small window that is divided. *R* represents the input resolutions. The specific formula for MHSA of the *n*-th window is as follows (Yuan et al. 2021a):

$$\text{MultiHead}(\mathbf{X}_n) = \text{Concat}\left[\text{Head}(\mathbf{X}_n)_1, \ldots, \text{Head}(\mathbf{X}_n)_H\right] \in \mathbb{R}^{K^2 \times C} \tag{2}$$

Where *K* denotes the size of each non-overlapping small window and *C* represents the number of channels.

$$\text{Head}(\mathbf{X}_n)_h = \text{Softmax}\left(\mathbf{X}_n \omega_m^h\right)\left(\mathbf{X}_n \omega_K^h\right)^T / \sqrt{C/H} \, \mathbf{X}_n \omega_v^h \in \mathbb{R}^{K^2 \times C/H} \tag{3}$$

$$\bar{\mathbf{X}}_n = \mathbf{X}_n + \text{MultiHead}(\mathbf{X}_n)\omega_o \in \mathbb{R}^{K^2 \times C/H} \tag{4}$$

$$\{\bar{\mathbf{X}}_1, \bar{\mathbf{X}}_2, \bar{\mathbf{X}}_3, \cdots, \bar{\mathbf{X}}_n\} \underset{\text{Merge}}{\mapsto} \mathbf{X}^{\text{MHSA}} \tag{5}$$

Where, $\omega$ denotes weight, $\omega_o \in \mathbb{R}^{C^2}$, $\omega_m^h \in \mathbb{R}^{C^2/H}$, $\omega_K^h \in \mathbb{R}^{C^2/H}$ and $\omega_v^h \in \mathbb{R}^{C^2/H}$ for $h \in \{1, \cdots, H\}$. *H* represents the number of heads. *C* represents the number of channels and $\bar{\mathbf{X}}_n$ represents the output representation of MHSA. $\mathbf{X}^{\text{MHSA}}$ denotes the features generated by the MHSA mechanism.

ΩSFormer changes the structure of the traditional Transformer between two MLPs in a feed-forward network by introducing a 3×3 deep convolution. Resolution Transformer (RT) (Fig. 4, left) aims to solve the problem of localized windowed self-attention performed separately on non-overlapping windows and lack of cross-window information exchange. Fig. 4 is modified from (Yuan et al. 2021a). This 3 × 3 deep convolution is implemented to leverage its dual advantages: enhancing localization and enabling cross-window interactions, thereby efficiently modeling non-overlapping localized windows. The deep convolution formula is:

$$\delta(\,) = \text{MLP}\{\text{DW\_Conv.}[\text{MLP}(\,)]\} \tag{6}$$

$\delta(\,)$ denotes the computational function of the nonlinear layer of the feed-forward network in the Transformer, DW_Conv. denotes the newly introduced 3 × 3 depth convolution and MLP indicates multi-layer perceptron.

Finally, the low-resolution features are upsampled to the highest resolution, and all cross-modal and cross-scale super-resolution features are combined to generate the output of ΩSFormer.

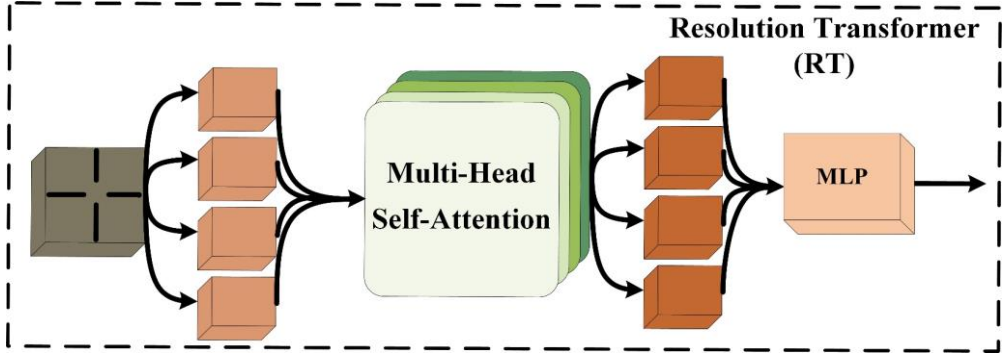

**Fig. 4.** The structure of resolution Transformer.

**3.3. Dual-modal Ω-like network**

ΩSFormer constructs a symmetric structure to fuse the rich high-level features of spectra and topography, thereby enhancing the semantic segmentation accuracy. In contrast to previous single resolution feature encoder and decoder structures, ΩSFormer maintains high-resolution features throughout each branch, while gradually downsampling to form a comprehensive feature fusion of low-resolution and high-resolution feature at each stage. As the code progresses from the lower to the upper layers, the number of scale features increases, resulting in the formation of the Ω-like structure with a wide top and a narrow bottom, as illustrated in Fig. 3. At the end of each stage, the cross-modal and cross-scale super-resolution features of the image and DEM modality are combined on the specified dimension, and the joint information of each modal is generated into a super-resolution feature through concatenation operation. ΩSFormer employs remotely sensed imagery to enhance the resolution of the DEM, thereby achieving the objective of improving labeling accuracy.

**3.4. Spatial topological semantic relationship optimization**

ΩSFormer utilizes the spatial connectivity of spatial topological semantic relationship (Fig. 5), in conjunction with the spatial distribution properties of pixels surrounding the critical edge. STSRO (Fig. 3, orange dashed box) uses soft segmentation to obtain initial attribute categories and combines topological relationships to optimize the accuracy of terraced field extraction. STSRO algorithm initially divides the spatial topological semantic relationship into a set of soft object regions, which represent the rough semantic segmentation result obtained from the backbone computation. Subsequently, the object region representation is calculated based on the preliminary semantic segmentation results (soft object regions) and the pixel feature representation of the deepest output of the network.

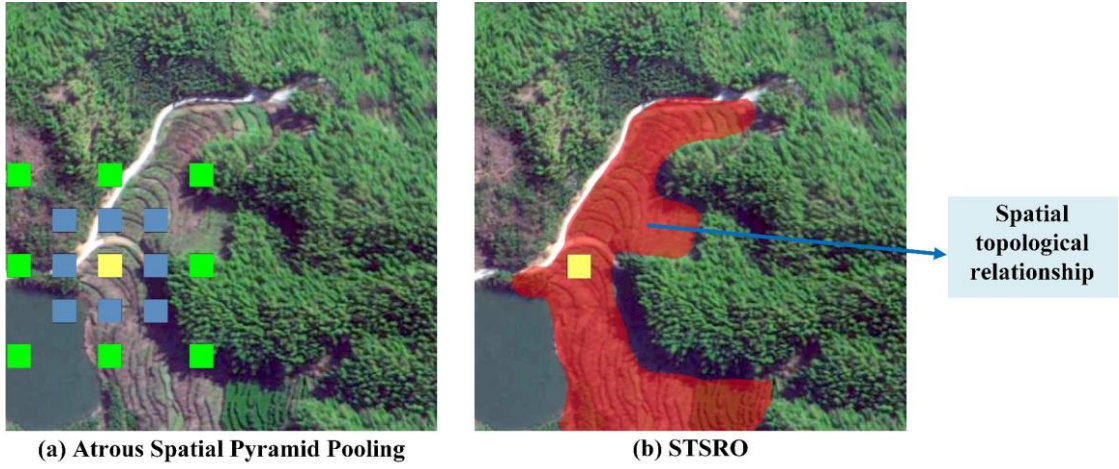

**Fig. 5.** Illustrating the STSRO spatial topological semantic relationship for the pixel marked with the yellow square. (a) Atrous spatial pyramid pooling: the multi-scale context is a set of sparsely sampled pixels marked with blue squares and green squares. The pixels exhibiting different colors are indicative of disparate dilation rates. These pixels are dispersed throughout both the object region and the background region. (b) STSRO: The spatial topological semantic relationship is expected to be a set of pixels lying in the object (marked with color red).

Representations for the *m*-th target region are obtained by aggregating their respective degrees of importance, thus forming a comprehensive representation of the *m*-th target region.

$$\boldsymbol{\rho}_m = \sum_{i \in E} \tilde{n}_{mi} \boldsymbol{\upsilon}_i \tag{7}$$

Here, $\boldsymbol{\rho}_m$ is the *m*-th object region representation, $\boldsymbol{\upsilon}_i$ is the representation of the *i*-th pixel and $E$ refers to the set of pixels in the image. $\tilde{n}_{mi}$ is the normalized degree for pixel belonging to the *m*-th object region.

Subsequently, pixel-region relation between the pixel representations of the deepest output of the DNN and the computed object region representation is computed. The object region representation is then weighted and summed based on the values of each pixel and the object region representation. Finally, the weighted sum of the object region representation is obtained by applying the value of each pixel and object region representation in the pixel-region relation matrix to the weighted sum of the object region representation.

$$\mu_{im} = e^{\lambda(\boldsymbol{\upsilon}_i, \boldsymbol{\rho}_l)} \Big/ \sum_{k=1}^{m} e^{\lambda(\boldsymbol{\upsilon}_i, \boldsymbol{\rho}_k)} \tag{8}$$

Here, $\lambda(\boldsymbol{\upsilon}, \boldsymbol{\rho}) = \chi(\boldsymbol{\upsilon})^{\mathrm{T}} \phi(\boldsymbol{\rho})$ is the unnormalized relation function. $\chi(\ )$ and $\phi(\ )$ are two transformation functions implemented by 1 × 1 conv and BN then ReLU, which is inspired by self-attention for a better relation estimation (Vaswani 2017). We rephrase the STSRO by using the Transformer encoder-decoder architecture in Fig. 6 (Yuan et al. 2021a).

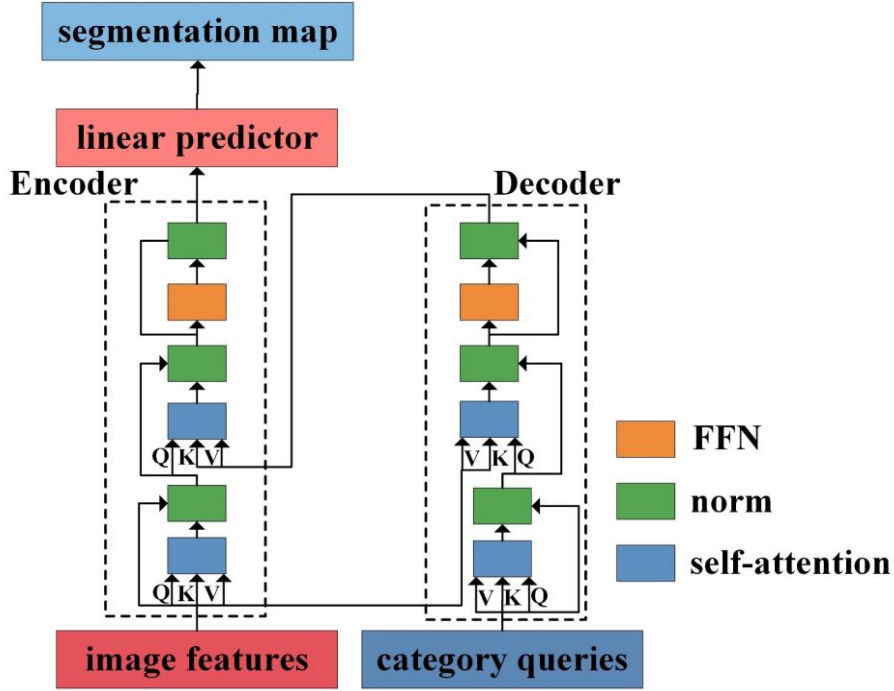

**Fig. 6.** The structure of segmentation Transformer.

Finally, the result of concatenating the spatial topological semantic relationship with the pixel representations from the deepest layer of the network is used as the augmented representation. ΩSFormer predicts the semantic category of each pixel based on the last spatial topological semantic relationship augmented representation with spatial dependency, which effectively improves the accuracy of the segmentation process.

**3.5. Vectorization extraction module**

VEM transforms contour vibration neural network into the image feature space. In VEM, the contour vibration equations are parameterized as $\beta(d), \mu(d)$, which changes dynamically with the input state $d$. Thus, the dynamic model is established as follows (Xu et al. 2022):

$$\frac{\partial^2 \boldsymbol{k}}{\partial t^2} - \beta(d)\frac{\partial^2 \boldsymbol{k}}{\partial d^2} + \mu(d)\frac{\partial \boldsymbol{k}}{\partial t} = 0 \qquad (9)$$

Where $d$ denotes the positions of contour string, $t$ denotes the time and $\boldsymbol{k}$ represents the displacement from equilibrium state (also real contour) at time $t$.

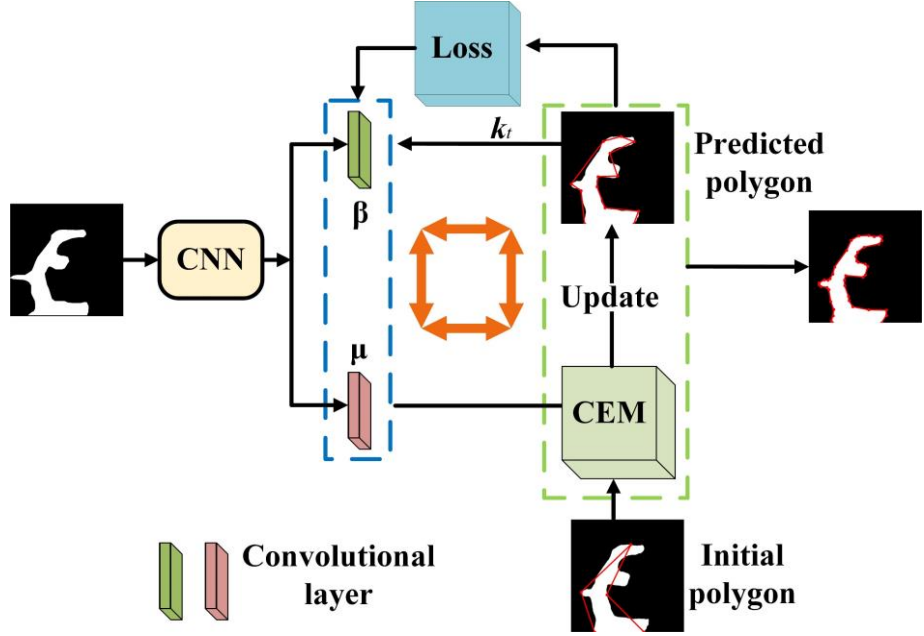

**Fig. 7.** An illustration of VEM.

The reparametrized equation coefficients are utilized as learnable states of the feature domain. The contour changes of the terraced field subsequently evolve in an asymptotic manner through the computation of the contour vibration equations (Xu et al. 2022), thereby enhancing the edge accuracy of the terraced field extraction. This method could effectively improve the lightweight and infinite scalability of terraced field extraction. The position of the target undergoes a transformation as a consequence of the combined action of internal and external forces. This can be conceptualized in image space as the influence of observed features (e.g., convolutional features), which exert a pulling or resisting force on contour changes. According to the discrete equations, contour vibration neural network can be expressed as follows:

$$\left[\bm{k}_{t+1}+\bm{k}_{t-1}-2\bm{k}_t\big/(\Delta t)^2\right]_o - \left[\beta_{o+1}\bm{k}_{t,o+1}-2\beta_o\bm{k}_{t,o}\right] + \mu_o\left[(\bm{k}_t-\bm{k}_{t-1})\big/\Delta t\right]_o = 0 \quad (10)$$

Where $o$ is one vertex of the contour, $\bm{k}_{t,o}$ denotes the contour point information of vertex $o$ at the time $t$. $\beta_o = [\beta(d)]_o$ and $\mu_o = [\mu(d)]_o$ with the abbreviation.

In order to detect the contour of terraced field, it is necessary to solve for $\bm{k}$ in the contour vibration neural network by Eq. (10). Consequently, the following text presents the contour evolution process along the time axis. For $x$ sampling points, the contour of the object is represented by a vector of $x$ vertices. The solution to the equation for $k_{t,x}$ is as follows:

$$\bm{k}_t = \left[k_{t,1}, k_{t,2}, \cdots, k_{t,x}\right]^{\mathrm{T}} \quad (11)$$

For *x* sampling points, the contour of the object is represented by a vector of *x* vertices.

The contour of the current time (*t*+1) can be calculated recursively from the state and estimated parameters of the previous time(*t*). The specific formula is as follows:

$$\boldsymbol{k}_{t+1} = 2\boldsymbol{k}_t - \boldsymbol{k}_{t+1} + \mathbf{B}\boldsymbol{k}_t(\Delta t)^2 - \mathbf{U} \odot (\boldsymbol{k}_t - \boldsymbol{k}_{t-1})\Delta t \tag{12}$$

Where $\mathbf{U} = [\mu_1, \mu_2, \cdots, \mu_x]^{\mathrm{T}}$, $\odot$ denotes the element-wise multiplication and the matrix **B** calculated based on the parameter update process in contour vibration neural network. The current contour $\boldsymbol{k}_t$ of the object allows for the direct indexing of the parameters $\beta(\boldsymbol{k}_t)$ and $\mu(\boldsymbol{k}_t)$ via the bilinear interpolation operation. To facilitate more rapid evolution, it is essential to commence with an optimal initial contour. We follow the same strategy (Cheng et al. 2019) to assign the initial object contour $\boldsymbol{k}_0$. Based on the estimated vibration parameters $\{\beta, \mu\}$ and the previous contours $\{\boldsymbol{k}_t, \boldsymbol{k}_{t+1}\}$, it is possible to infer the next contour $\boldsymbol{k}_{t+1}$. Finally, the object contours are refined through an iterative evolution process.

### 3.6. Loss functions

To improve the accuracy of semantic segmentation for target edges in TFVE and address potential imbalances in category distributions during sample screening, a loss function was developed as a weighted sum of three distinct loss components.

BCEWithLogitsLoss is an adaptation of the standard binary cross-entropy loss function, designed to enhance the handling of imbalanced datasets. It is designed to assign appropriate weights to samples in each category, thereby ensuring a balanced distribution across different categories.

$$L_1 = -\frac{1}{C}\sum_{x=1}^{C}\left[\omega_{\text{pos}} y_x \log(p_x) + \omega_{\text{neg}}(1-y_x)\log(1-p_x)\right] \tag{13}$$

Where *C* represents the total number of samples, $y_x$ denotes the true label of the *x*-th sample, $p_x$ is the probability that the model predicts the *x*-th sample to be a positive example, $\omega_{\text{pos}}$ is the weight of the positive samples and $\omega_{\text{neg}}$ is the weight of the negative samples.

Lovasz-softmax loss (Berman et al. 2018) is utilized to smooth the boundaries of the segmentation result, thereby generating a more continuous prediction boundary. Since TFVE is a binary classification problem, the indexing of foreground categories (i.e., terraced field) will be considered to be optimized by using a max margin classifier.

$$\mathbf{A}_c(\boldsymbol{y}_t, \boldsymbol{y}_o) = \{\boldsymbol{y}_t \neq c, \boldsymbol{y}_o = c\} \cup \{\boldsymbol{y}_t = c, \boldsymbol{y}_o \neq c\} \tag{14}$$

Where $\boldsymbol{y}_o$ represents the semantic segmentation output, $\boldsymbol{y}_t$ represents the ground truth and $c$ denotes the set of mispredicted pixels for class $c$.

$$\Delta_{Jc} : \mathbf{A}_c \in \{0,1\}^f \mapsto |\mathbf{A}_c| / |\{\mathbf{y}_t = c\} \cup \mathbf{A}_c| \tag{15}$$

Where $f$ is the number of pixels in the image or minibatch considered. Subsets of pixels with their indicator vector in the discrete hypercube are identified. The formula is as follows:

$$L_2 = \overline{\Delta}_{J1}(a(S)) \tag{16}$$

With $\overline{\Delta}_{J1}$ the lovasz extension to $\Delta_{Jc}$, the resulting loss surrogate is the lovasz hinge applied to the Jaccard loss. Where the vector of hinge losses $a \in \mathbb{R}^+$ is the vectors of errors discussed and $S$ represents the output scores.

In VEM, the symmetric chamfer distance is employed as the loss function to match the estimated polygon $\overline{Y}$ with the ground truth polygon $Y$. Both $\overline{Y}$ and $Y$ are the coordinate sets of the contour vertices. The formula is as follows:

$$L_3(\overline{Y}, Y) \approx \sum_{\overline{k} \in \overline{Y}} \min_{k \in Y} \|\overline{k} - k\|_2^2 + \sum_{k \in Y} \min_{\overline{k} \in \overline{Y}} \|k - \overline{k}\|_2^2 \tag{17}$$

Where, $k$ denotes the contour point information of vertex.

The final loss value is computed as a weighted sum of $L_1, L_2$ and $L_3$.

$$L = L_1 + \gamma L_2 + \theta L_3 \tag{18}$$

Where $\gamma$ and $\theta$ denote weights that are hyperparameters.

**4 Results and discussion**

In this section, we evaluate the performance and generalizability of ΩSFormer in comparison to state-of-the-art (SOTA) networks using metrics such as mIoU and overall accuracy (OA). We present the extraction results of ΩSFormer alongside those of the comparative networks, such as U-Net (Ronneberger et al. 2015), U-Net++ (Zhou et al. 2020b), DeepLabV3+ (Chen et al. 2018a), SegFormer (Yuan et al. 2021a), HRNet (Ke et al. 2019), HRFormer (Yuan et al. 2021a) W-Net (Xia and Kulis 2017), W-Net++ (Limei et al. 2021) and SAM (Kirillov et al. 2023),.

Training Details: ΩSFormer was trained using distributed training across four NVIDIA GeForce RTX 3090 Ti GPUs with 24 GB of memory each. ΩSFormer was designed to operate on the Windows operating system and utilizes the PyTorch with the AdamW optimizer. The initial learning rate was set to $1 \times 10^{-3}$ and adjusted dynamically using the Cosine Annealing learning rate strategy. Weight decay was set to $1 \times 10^{-4}$, and dropout rates ranged from 0.6 to 0.8.

## 4.1 Comparison among different modals and different branches

Table 2. Summary of results for comparison networks.

| single-modal | single-branch dual-modal | dual-branch dual-modal | Net | Input | mIoU | OA |
|---|---|---|---|---|---|---|
| √ | × | × | U-Net | RGB | 0.718 | 0.776 |
| √ | × | × | U-Net | DEM | 0.518 | 0.568 |
| √ | × | × | U-Net++ | RGB | 0.766 | 0.812 |
| √ | × | × | U-Net++ | DEM | 0.615 | 0.639 |
| √ | × | × | DeepLabV3+ | RGB | 0.762 | 0.815 |
| √ | × | × | DeepLabV3+ | DEM | 0.492 | 0.571 |
| √ | × | × | SegFormer | RGB | 0.791 | 0.829 |
| √ | × | × | SegFormer | DEM | 0.581 | 0.568 |
| √ | × | × | HRNet | RGB | 0.792 | 0.831 |
| √ | × | × | HRNet | DEM | 0.618 | 0.642 |
| √ | × | × | HRFormer | RGB | 0.811 | 0.837 |
| √ | × | × | HRFormer | DEM | 0.679 | 0.692 |
| √ | × | × | SAM | RGB | 0.794 | 0.849 |
| √ | × | × | SAM | DEM | 0.611 | 0.602 |
| × | √ | × | DeepLabV3+ | RGB+DEM | 0.808 | 0.831 |
| × | √ | × | SegFormer | RGB+DEM | 0.812 | 0.843 |
| × | √ | × | HRNet | RGB+DEM | 0.821 | 0.854 |
| × | √ | × | HRFormer | RGB+DEM | 0.846 | 0.871 |
| × | × | √ | W-Net | RGB+DEM | 0.846 | 0.881 |
| × | × | √ | W-Net++ | RGB+DEM | 0.848 | 0.893 |
| × | × | √ | ΩSFormer | RGB+DEM | **0.976** | **0.981** |

The dual-modal approach allows for the comprehensive consideration of the spectral characteristics of remotely sensed imagery and the surface morphology features present in DEM. This makes it a more suitable method for the extraction of features in terraced field with a stepped distribution. Therefore, single-modal SOTA networks, single-branch and dual-branch dual-modal SOTA networks are compared and experimented. The comparison of the experimental results is as follows:

(1) Comparison between single-modal and dual-modal

Results presented in Table 2 show that, among experiments with single-modal inputs, the outcomes achieved using remotely sensed imagery surpass those obtained with DEM input. The comparison results between single-modal and dual-modal input methods indicate that, the D$^2$MB input method, designed by ΩSFormer, incorporating dual-modal super-resolution features, demonstrated the superior performance in extracting finer semantic information. ΩSFormer achieved the highest accuracy and exhibited a significant advantage in TFVE. The evaluation of ΩSFormer with DMRVD resulted in a mIoU of 0.976. The result is 0.165 higher than the optimal single-modal results achieved using remotely sensed imagery and 0.297 higher than the best obtained using DEM. Compared with other dual-modal approaches, the highest improvement achieved was 0.168, while the lowest was 0.128. Fig. 8 shows that the semantic segmentation results generated by ΩSFormer closely resemble the original images and labels. ΩSFormer significantly improves edge segmentation accuracy, capturing target edges more precisely and producing finer segmentation results.

(2) Comparison between single-branch dual-modal and dual-branch dual-modal

Although dual-modal input methods can extract rich and useful information and enhance features, different dual-modal input methods (single or dual-branch) can also affect semantic segmentation accuracy. In order to investigate the impact of dual-modal input method on TFVE, single-branch and dual-branch dual-modal input methods are compared in experiments. The results of this comparison are presented in Table 2.

In this study, remotely sensed imagery and DEM were integrated into a four-channel format before being input into the network. The combined four-channel dataset was processed through the network using a single branch. The fusion of remotely sensed imagery and DEM were demonstrated prior to input, utilizing a single-branch input SOTA networks that incorporates four channels of data.

D$^2$MB performs better than the single-branch dual-modal input method in TFVE. The mIoU of ΩSFormer is demonstrably superior to that of a single-branch mIoU. It achieves an mIoU that is 0.130 higher than the optimal mIoU obtained with single-branch input. In comparison to a single-branch dual-modal input approach, D$^2$MB facilitates the extraction of distinctive semantic information from multiple images, which is then synthesized into a cohesive and refined feature representation. In addition to reducing the incidence of registration errors and other issues that may arise during the preprocessing of images, this approach also enhances the accuracy of TFVE by leveraging multiple sources of information. The extraction visualization results in Fig. 8 show that the extraction with $D^2MB$ outperforms the single-branch input method with channel fusion and achieves better results on the TFVE.

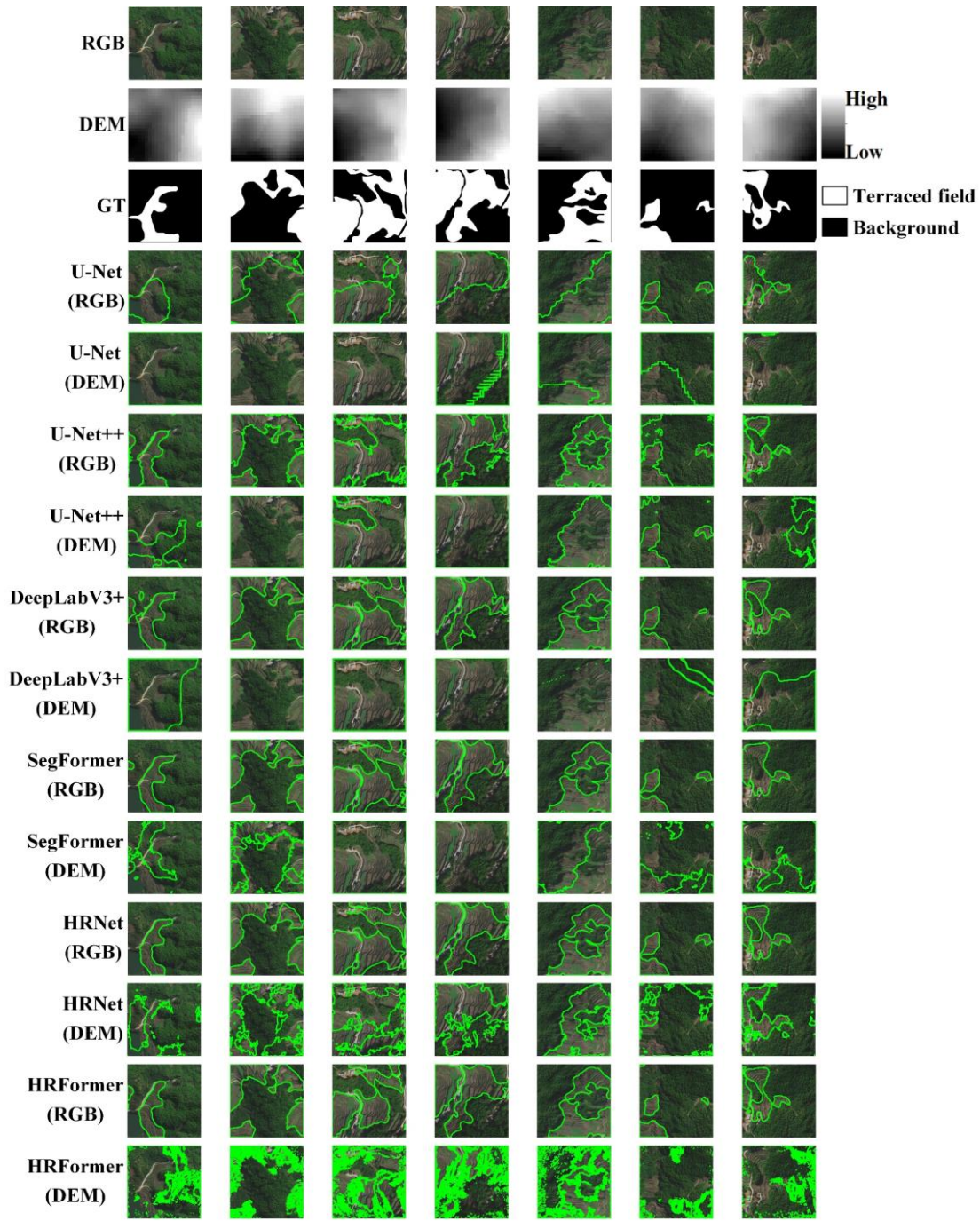

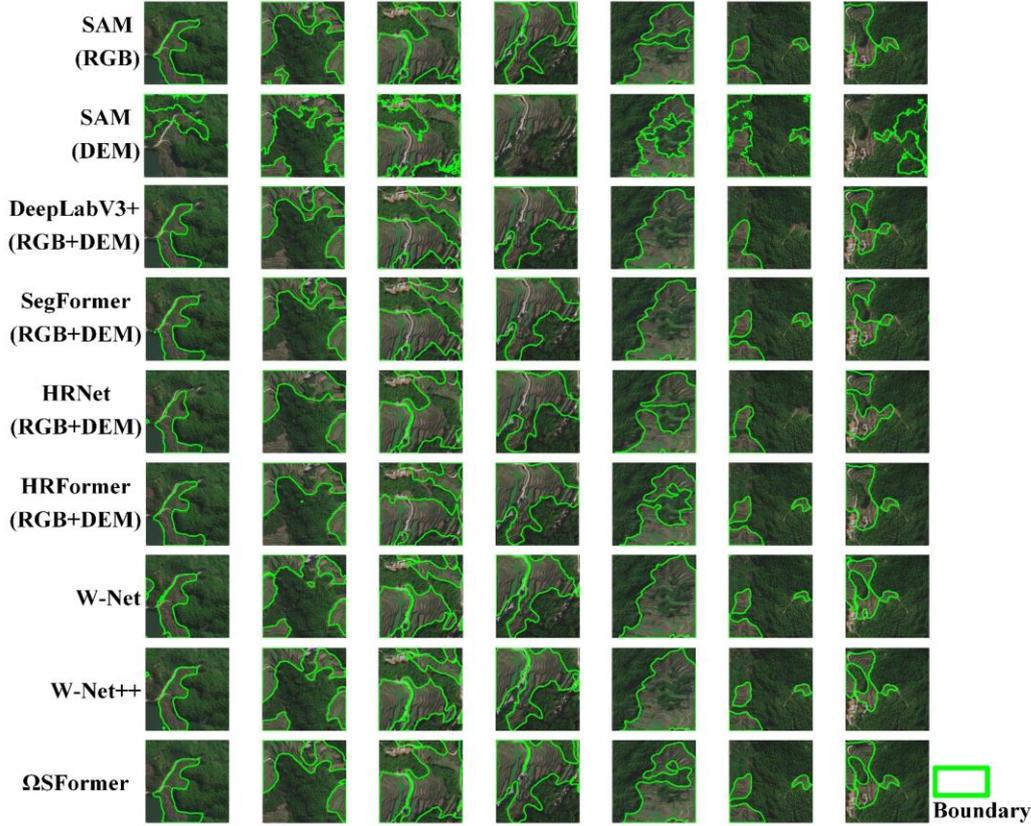

**Fig. 8.** Visualization results of DMRVD compared with comparison networks.

### 4.2 Ablation experiment

To verify the impact and effectiveness of STSRO on semantic segmentation results, an ablation experiment was designed to verify the effectiveness of the module. The results show (Fig. 9, Table 3) that the inclusion of STSRO can enhance semantic segmentation accuracy to a certain extent.

Table 3. Summary of results for ablation experiment.

|  | mIoU | OA |
|---|---|---|
| STSRO | 0.976 | 0.981 |
| No STSRO | 0.941 | 0.953 |

As illustrated in Table 3, the experimental results show that after incorporating STSRO, the mIoU and OA increased by 0.035 and 0.028, respectively. Consequently, STSRO enhances the accuracy of terraced field delineation and refines edge extraction results for terraced fields.

### 4.3 Cross-scale experiments

Due to significant differences in capturing object features and details across varying resolutions, there is an imbalance in ability of the model to perceive information from different branches. Cross-scale inputs involve the model receiving imagery from different resolutions simultaneously, which may introduce issues of scale inconsistency and hinder ability of the model to effectively learn and extract object features. Considering the scale effects caused by cross-modal differences, resampling techniques are employed to standardize the resolution of three data types: images, labels and DEM. This is done in order to verify the optimal fusion scheme of super-resolution across modal scales (remotely sensed imagery and DEM spatial resolution inconsistency). The resulting comparative datasets (Table 1) are subsequently utilized for comparative experiments.

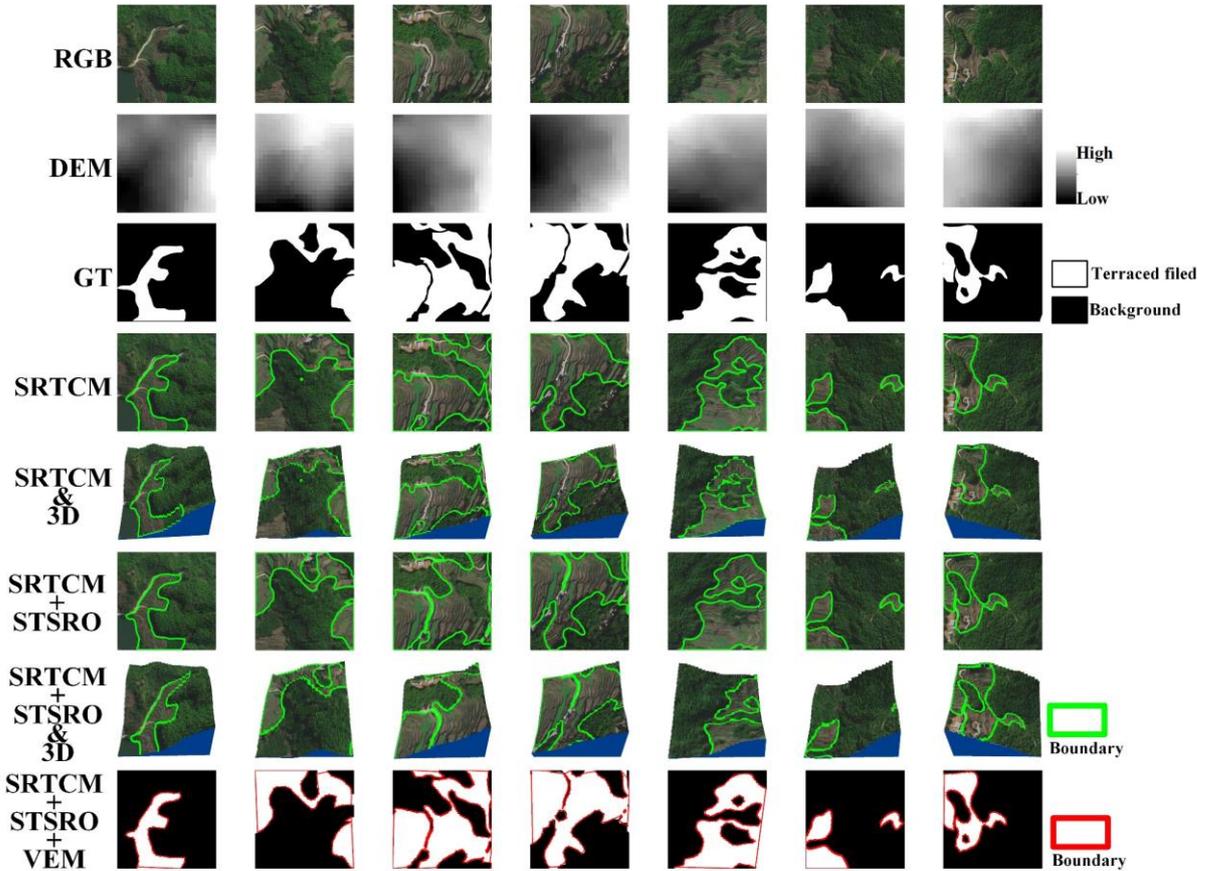

**Fig. 9.** Comparative visualization of ablation experiments. SRTCM indicate super-resolution Transformer connection module, STSRO denotes spatial topological semantic relationship optimization.

Table 4. Summary of results for cross-scale effect.

| RGB | DEM | Label | mIoU | OA |
|---|---|---|---|---|
| 2m | 12.5m | 2m | 0.883 | 0.905 |
| 2m | 2m | 2m | 0.976 | 0.981 |
| 12.5m | 12.5m | 12.5m | 0.921 | 0.947 |

Table 4 shows that the semantic segmentation results of cross-scale are the worst. The mIoU is 0.883, which is 0.093 lower than the dual-modal data results of the same resolution. In contrast, within the experiments at the dual-modal same-scale, the mIoU for the 2m resolution is 0.976, whereas that for the 12.5m resolution is 0.921. Fig. 10 illustrates that the semantic segmentation accuracy can be enhanced by employing DMRVD at the same scale. Furthermore, it can be observed that semantic segmentation accuracy increases with an increase in image resolution.

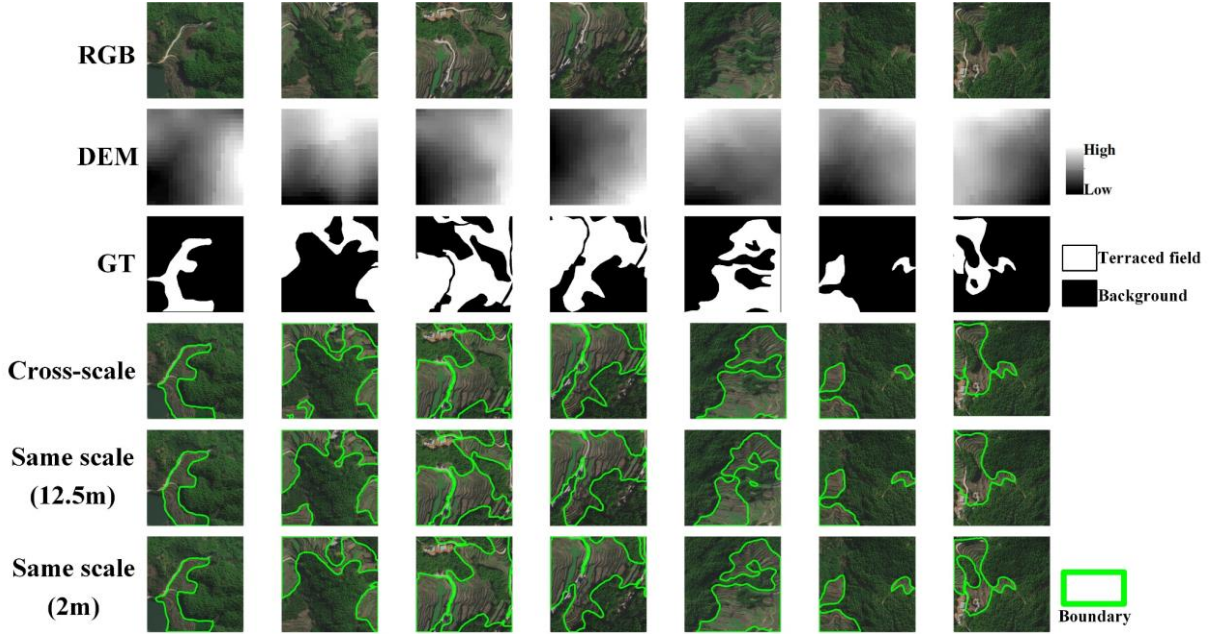

**Fig. 10.** Experimental results of cross-modal comparison on dual-modal datasets.

**4.4 Discussion**

4.4.1 The quality of DMRVD

A combination of classical DNNs and SOTA networks are employed to conduct comprehensive experiments that demonstrate the effectiveness and superiority of DMRVD. Fig. 8 demonstrates that DMRVD achieves favorable results on existing SOTA networks, with the delineated terraced field boundaries closely aligning with the ground truth labels. This suggests that the vector boundaries in DMRVD are precise and of high quality. Commonly, the quality of a dataset can be evaluated based on the degree of accuracy achieved in the classification process (Hong et al. 2023; Tiwari et al. 2023). Table 2 shows the semantic segmentation accuracy, indicating that DMRVD performs better overall compared to SOTA networks. For DMRVD, ΩSFormer achieves the highest accuracy, while U-Net exhibits the lowest accuracy. This suggests that DMRVD is more applicable and of superior quality.

DMRVD covers four provinces in China with significant differences in terrain and landforms, resulting in relatively weak spatial autocorrelation, making it representative and universal. The information present in remotely sensed imagery is insufficient for capturing the distinctive distribution characteristics of terraced field. In contrast, DMRVD comprises multimodal data, including remotely sensed imagery, DEM and vector data. This integration implies that DMRVD encompasses advanced and diverse features, thereby facilitating comprehensive updates to land cover. This process helps to mitigate the impact of DNN learning errors caused by noise from misinterpreted labels, thereby enhancing the accuracy of terraced field edge extraction. Terraced field extraction is both challenging and significant. However, the lack of deep-learning-based multimodal datasets has precluded a comprehensive investigation to date. The construction of a multimodal raster-vector dataset hinges upon the assurance of both the quantity and quality of the data. The DMRVD, which is of overall good quality, makes a significant contribution to the process of extracting terraced field, SWC and agricultural applications (Weiss et al. 2020).

4.4.2 The effectiveness of SRTCM

To enhance semantic segmentation accuracy, SRTCM was proposed for the first time. Experimental comparisons have demonstrated the efficacy of SRTCM, as evidenced by the results in Table 2, where $\Omega$SFormer achieved the highest accuracy. Fig. 8 illustrates that the edge semantic segmentation results of the $\Omega$SFormer are more closely aligned with the ground truth than those of other SOTA networks. This is due to the fact that the SRTCM integrates the original high-resolution features from the dual-modal data with the mid- and low-resolution features at each step of encoder, utilizing a MHSA mechanism. The issue of spatial dimension reduction in input feature maps resulting from classical encoder and downsampling inevitably results in the loss of detailed information. The retention of the original resolution features serves to overcome this problem by the preservation of more precise detail information. Consequently, SRTCM ensures that the details that have been lost can be better recovered during the decoder phase, thereby improving the accuracy of edge semantic segmentation. The application of multi-scale feature fusion for various computer vision tasks has the potential to yield enhanced results (Ke et al. 2019). SRTCM proposes a cross-modal and cross-scale super-resolution feature fusion approach, which combines the distinctive characteristics of different resolutions. This methodology effectively preserves semantic information of the original resolution while enhancing the accuracy of semantic segmentation.

4.4.3 The performance of D$^2$MB

Comparative experiments were conducted to verify the effectiveness of D$^2$MB. Table 2 indicates that the accuracy of D$^2$MB is superior to that of single-modal and single-branch dual-modal. Fig. 8 also illustrates that the D$^2$MB structure of ΩSFormer exhibits a notable advantage in the terraced field extraction task, with a higher degree of extraction accuracy. The enhanced accuracy of dual-branch over single-branch can be attributed to that (1) the input of dual-modal data can derive various advanced and rich features; (2) the integration of dual-modal features derived from remotely sensed imagery and DEM not only captures the spectral characteristics of objects but also incorporates terrain morphology features, rendering it more suitable for the distinctive spatial distribution characteristics of terraced field. Both Tang et al. (2023) and Zhang et al. (2021a) have demonstrated that the results obtained by the dual-modal input methods are complete and well-bounded.

The reasons for D$^2$MB structure superior accuracy compared with single-branch dual-modal are that: (1) alignment errors in multimodal images can lead to image distortion and incorrect feature extraction. The single-branch structure is incapable of rectifying the alignment outcomes for remotely sensed imagery and DEM. Whereas D$^2$MB decodes-encodes individually and then fuses each modality. This approach offers the potential to rectify any errors that may arise during the multimodal alignment process, thereby enhancing the overall outcome of the multimodal alignment; (2) the selection of features in the single-branch structure is constrained by a single modality. D$^2$MB is aided by the integration of remotely sensed imagery and DEM in feature selection, with the resulting features then fused based on their spatial relationship. The comprehensive selection of semantic features is more conducive to improving the accuracy of the semantic segmentation; (3) D$^2$MB proposes that high-resolution remotely sensed imagery fuses low-resolution DEM to obtain the super-resolution DEM, thereby enhancing the accuracy of semantic segmentation. But single-modal structures are unable to obtain super-resolution images, the semantic information extracted is limited by the resolution of the available images; (4) the semantic information that can be obtained through single-modal is inherently limited. In contrast, D$^2$MB offers a means of combining and optimizing the spectral information and surface morphology features through the fusion of dual-modal data. This fusion allows for the extraction of optimized features from both modalities, thereby ensuring the accuracy of terraced field extraction.

4.4.4 The benefits of super-resolution fusion for enhancing accuracy

This paper processes the DEM to capture finer variations and features of terrain by enhancing the resolution of the DEM. Based on the results in Table 2 and Fig. 8, the details of low-resolution DEM is enhanced. The experimental results after super-resolution are significantly better than the semantic segmentation results using single-modal with DEM, effectively improving the edge accuracy of terraced field extraction. This is achieved by leveraging the power of cross-scale internal recursion in images to infer intricate and high-resolution image-specific relationships from DEM of LR image. This enables the LR image to recover more detailed texture, thereby enhancing its resolution (Shocher et al. 2018; Zhou et al. 2020a). ΩSFormer employs remotely sensed imagery to enhance the resolution of DEM, thereby improving the accuracy of label. Following super-resolution of ΩSFormer, the quality of the semantic segmentation result is enhanced, with the objective of ensuring that high-frequency information is not compromised. ΩSFormer ensures that each layer of the network integrates with high-resolution features, thereby preserving high-frequency information and enabling precise edge extraction results. This process alleviates issues such as unclear edges and poor performance that may arise after super-resolution.

4.4.5 The advantages of STSRO

ΩSFormer leveraged STSRO enhances target feature representation by calculating the spatial topological semantic relationship between pixels and multiple target regions. To facilitate comparison, an ablation experiment was conducted to assess the effectiveness of STSRO. The results show (Fig. 8, Table 3) that the inclusion of STSRO can enhance semantic segmentation accuracy to a certain extent. Semantic segmentation is the task of assigning a category label to each pixel in an image, aiming to achieve more accurate segmentation results. In recent years some scholars, such as Yuan et al. (2021b) and Xie et al. (2021) consider the relations between a position and its contextual positions. The reason STSRO enhances segmentation accuracy lies in its use of self-attention mechanisms based on the initial segmentation results. Each pixel performs similarity matching for different target categories, learning the relationships between each pixel and all category regions. Thereby enhancing ability of each pixel to represent its corresponding category region. Utilizing the representation of the corresponding object class to describe a pixel involves semantic segmentation based on the relationship between the pixel and the target class, rather than the relationship between individual pixels. Enhancement of semantic segmentation accuracy through spatial topological semantic relationships.

4.4.6 The effectiveness of VEM

In this work, VEM was incorporated after semantic segmentation to further vectorize the semantic segmentation results. According to the visualizations (Fig. 9), it can be observed that after VEM, the network becomes easier to handle the topological changes of terraced fields and more effectively capture sharp edges. VEM facilitates the prediction of effective boundaries of terraced field and generates more accurate predictions within the DMRVD. VEM utilizes a contour vibration neural network to perform multi-branch parameter optimization on the binarized results of STSRO. By leveraging a contour vibration neural network to the binarized semantic segmentation feature space, the results are made to more closely align with the ground truth. VEM reduces the inherent pixelation of raster images, thereby enhancing the clarity and smoothness of the target contours, and obtaining infinite scalability and lightweight results. Xu et al. (2022) and Zhang et al. (2018) have also demonstrated that vectorization offers significant advantages in object extraction tasks. Vectorized extraction is also employed for the purpose of data compression and to facilitate complex spatial analyses. VEM effectively improves the semantic segmentation of ΩSFormer.

**5 Conclusion**

In this study, in order to improve the accuracy of edge segmentation and to solve the problem that multi-source information cannot be effectively utilized, a novel semantic segmentation network for TFVE named ΩSFormer was proposed. This article makes the following research:

(1) We are first to establish DMRVD for deep-learning-based TFVE and validate its effectiveness and advantages in SOTA networks.

(2) This study proposes the ΩSFormer for the first time. It has been verified in a series of comparative and ablation experiments that it has the highest extraction accuracy and can obtain vectorized extraction results, providing a new idea and method for multi-modal high-accuracy remote sensing land feature extraction.

(3) This study is of great significance for the detection and automatic extraction of

terraced fields, which serves to monitoring soil erosion for SWC.

In future work, we will consider integrating more advanced networks, such as Mamba, which may offer both efficiency and high-accuracy extraction of terraced fields for SWC and other environmental monitoring.

**Acknowledgements**

The authors are grateful for the comments and contributions of the editors, anonymous reviewers and the members of the editorial team. This work was supported by the Key Program of the National Natural Science Foundation of China under Grant Nos. 42030102, National Natural Science Foundation of China (NSFC) under Grant Nos. 41771493 and 41101407, and the Fundamental Research Funds for the Central Universities under Grant CCNU22QN019.